\documentclass[11pt]{article}

\usepackage[final]{acl}
\usepackage{tcolorbox}
\usepackage[ruled,vlined]{algorithm2e}
\usepackage{times}
\usepackage{latexsym}

\usepackage{xcolor}
\usepackage{mdframed}
\usepackage{listings}

\usepackage{booktabs}
\usepackage{booktabs}
\usepackage{multirow}
\usepackage{makecell}

\usepackage{hyperref}
\usepackage{url}
\usepackage{graphicx}
\usepackage{amsmath} 
\usepackage{multirow}
\usepackage{authblk}
\usepackage[T1]{fontenc}

\usepackage[utf8]{inputenc}

\usepackage{microtype}

\usepackage{inconsolata}

\usepackage{graphicx}

\title{TableCache: Primary Foreign Key Guided KV Cache Precomputation for Low Latency Text-to-SQL}

\author[1,2]{\textbf{Jinbo Su}}
\author[1,2]{\textbf{Yuxuan Hu}}
\author[1,3]{\textbf{Cuiping Li}}
\author[1,3]{\textbf{Hong Chen}}
\author[4]{\textbf{Jia Li}}
\author[4]{\textbf{Lintao Ma}}
\author[1,2]{\\\textbf{Jing Zhang}\thanks{Corresponding author}}

\affil[1]{School of Information, Renmin University of China,Beijing, China}
\affil[2]{Key Laboratory of Data Engineering and Knowledge Engineering, Beijing, China}
\affil[3]{Engineering Research Center of Database and Business Intelligence, Beijing, China}
\affil[4]{Ant Group, Hangzhou, China}

\affil[ ]{ \textit{\{sujinbo,zhang-jing\}@ruc.edu.cn}}

\begin{document}
\maketitle
\begin{abstract}
In Text-to-SQL tasks, existing LLM-based methods often include extensive database schemas in prompts, leading to long context lengths and increased prefilling latency. While user queries typically focus on recurrent table sets—offering an opportunity for KV cache sharing across queries—current inference engines, such as SGLang and vLLM, generate redundant prefix cache copies when processing user queries with varying table orders.
To address this inefficiency, we propose precomputing table representations as KV caches offline and querying the required ones online. A key aspect of our approach is the computation of table caches while preserving primary foreign key relationships between tables. Additionally, we construct a Table Trie structure to facilitate efficient KV cache lookups during inference.
To enhance cache performance, we introduce a cache management system with a query reranking strategy to improve cache hit rates and a computation loading pipeline for parallelizing model inference and cache loading.
Experimental results show that our proposed TableCache achieves up to a 3.62× speedup in Time to First Token (TTFT) with negligible performance degradation.
\end{abstract}

\section{Introduction}
The Text-to-SQL task endeavors to translate natural language (NL) queries into precise structured query language (SQL) statements, enabling efficient database execution. By providing a natural language interface, it empowers non-technical users to interact with databases effortlessly, unlocking seamless access to targeted data. Recently, this task has garnered growing interest from both the NLP and database research communities.


Modern Text-to-SQL approaches primarily utilize LLM-based multi-agent systems, where agents process user queries and database schemas (i.e., table metadata) as input, coordinating a series of complex steps to generate the desired SQL statement.
Although current LLM-based agents achieve state-of-the-art performance on various benchmarks, they face significant deployment challenges. A major bottleneck arises from embedding extensive table metadata into prompts, leading to longer sequences and substantially increased prefill latency.



\begin{figure}
  \centering
  \includegraphics[width=1\linewidth]{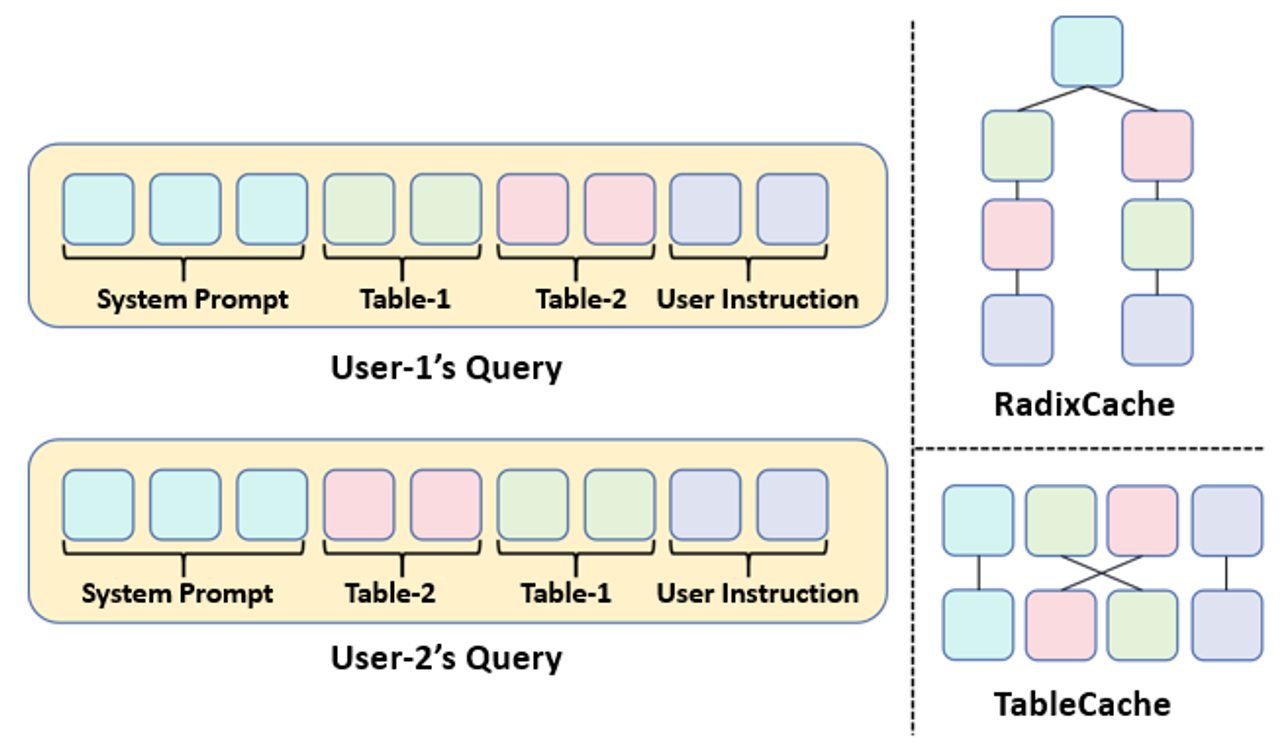}
  \caption{An example of TableCache}
  \label{fig:case}
\end{figure}

To address the high latency commonly associated with LLM inference, advanced inference engines like SGLang~\citep{zheng2023efficiently} and vLLM~\citep{kwon2023efficient} have become essential tools for improving inference efficiency. These systems enhance LLM deployment mainly through efficient KV cache management and have shown strong performance across a variety of tasks. A key mechanism they rely on is the use of structures such as RadixCache or PrefixCache, which organize previously computed KV cache to enable reuse—though this requires exact prefix matches. However, as shown in Figure~\ref{fig:case}, the dynamic ordering of tables in user queries often leads to prefix matching failures, compelling the system to produce multiple independent KV cache paths and thereby reducing cache efficiency. Another limitation is that this strategy does not fully leverage distinctive attributes of Text-to-SQL workloads: while the static structure of database schemas allows for table-level KV cache reuse, query trends often generate localized data hotspots. Together, these characteristics highlight the potential to accelerate the LLM prefill phase by exploiting shared table information.

Based on these observations, we propose \textbf{TableCache}, a method to accelerate LLM inference for Text-to-SQL tasks by precomputing KV caches for tables offline. Specifically, we proactively identify all tables in the database schemas that users may query, represent these tables as KV cache values offline, and store the results in CPU memory. The core idea is to represent tables while preserving primary foreign key relationships, preventing performance degradation caused by the loss of cross-attention when encoding tables individually. To achieve this, we construct a primary foreign key graph from the database, apply topological sorting to determine the sequence of tables, and use causal attention based encoding with LLMs to represent the tables. This approach restores essential relationships between tables, maintaining both efficiency and system performance.

To streamline the online query process, all tables are organized into a trie structure, referred to as the \textbf{Table Trie}. When a user query arrives, the system iteratively matches table positions within the Table Trie. Upon finding an exact match for a table, the corresponding KV cache is loaded from CPU memory to GPU memory. The offline KV caches are then concatenated and passed to the LLM for SQL generation.

To further optimize cache performance, we introduce a cache management system that implements CPU-GPU cache eviction and loading policies, such as FIFO and LRU. More importantly, we propose two strategies to enhance system efficiency: (1) a \textbf{query reranking strategy} that improves cache hit rates by leveraging user query similarities, and (2) a \textbf{computation loading pipeline} that parallelizes model inference with cache loading by asynchronously transferring the KV cache required for subsequent user queries from CPU to GPU during the computation of the current query.


In summary, the contributions of this paper are delineated as follows:

\begin{itemize}

\item We propose TableCache, a method that recomputes KV caches for tables while preserving primary foreign key relationships and builds a Table Trie for efficient table matching, ensuring high performance and efficiency for Text-to-SQL LLMs.

\item We enhance cache performance with a query reranking strategy to improve hit rates and a computation loading pipeline to parallelize model inference and cache loading.

\item Experimental results demonstrate that our proposed TableCache achieves a speedup of up to 3.62$\times$ in TTFT while maintaining the original model's performance.
\end{itemize}

\section{Related Work}
\subsection{KV Cache Management}





On the system side, KV cache optimization focuses on memory management and scheduling. Both vLLM and SGLang employ PrefixCache to reuse shared prefixes across sequences.

For retrieval-augmented generation(RAG) scenarios, methods like PromptCache~\citep{Gim2023PromptCM}, TurboRAG~\citep{lu2024turborag}, and BlockAttention~\citep{ma2024block} utilize block-based pre-encoding but neglect inter-block attention relationships, causing performance degradation. In response, CacheBlend~\citep{yao2024cacheblendfastlargelanguage}, KVLink~\citep{yang2025kvlink}, KVZip~\citep{kim2025kvzip}, and EPIC~\citep{hu2024epic} introduce selective recomputation for more accurate cache reuse. Unlike RAG, Text-to-SQL provides explicit attention reconstruction guidance through primary foreign key, inspiring our primary foreign key guided attention reconstruction mechanism.


On the algorithmic side, research focuses on eliminating redundant or less useful KV cache entries to improve efficiency. Methods such as H2O~\citep{Zhang2023H2OHO}, QUEST~\citep{tang2024quest}, Ada-KV~\citep{feng2024ada}, snapKV~\citep{li2024snapkv}, and PyramidKV~\citep{Cai2024PyramidKVDK} have designed various KV cache eviction and compensation schemes to accelerate inference. In contrast to them, our TableCache preserves all KV caches to suit the high-reuse setting of Text-to-SQL.

\begin{figure*}
  \centering
  \includegraphics[width=1\linewidth]{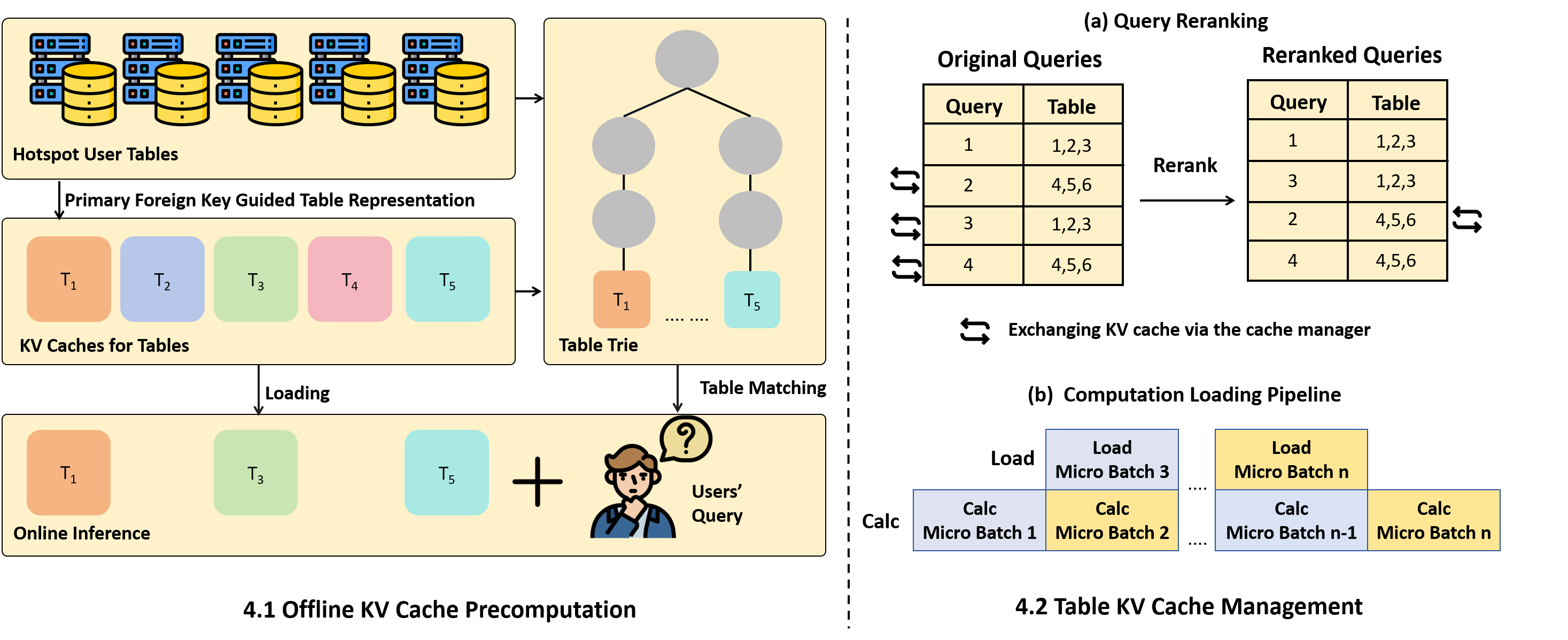}
  \caption{Overview of TableCache. KV caches are precomputed and organized into a hierarchical Table Trie during offline preprocessing. Inference utilizes cache management system, including query reranking for cache-friendly sequencing and a computation loading pipeline to reduce latency and improve efficiency.}
  \label{fig:main}
\end{figure*}

\subsection{Text-to-SQL}
Mainstream LLM-based approaches for Text-to-SQL fall into two paradigms: prompt engineering and supervised fine-tuning (SFT).
Prompt engineering methods often leverage few-shot prompting strategies~\citep{cheng2023binding,nan2023enhancing}. Examples include DIN-SQL\citep{pourreza2023din} and CHESS\citep{talaei2024chess}, which decompose the task into subtasks like Schema Linking and SQL Generation with a self-correction mechanism.
In contrast, SFT-based approaches~\citep{gao2024xiyan,pourreza2024dts}, such as CODES~\citep{li2024codes}, DAIL-SQL~\citep{gao2024text}, and OmniSQL~\citep{li2025omnisql}, focus on fine-tuning the end-to-end Text-to-SQL capabilities of LLMs. Unlike prior work, our research focuses on enhancing the inference speed of Text-to-SQL models in practical deployment scenarios where they serve end users. 



\section{Problem Setup}

The Text-to-SQL problem can be formally defined as follows: given a set of databases, each consisting of multiple tables, a natural language query \(q\) issued on a database \(D\) requires retrieving a collection of \(m\) tables \(T = \{T_{1}, ..., T_{m}\}\) to answer \(q\). These tables may include all tables in \(D\) or a subset pre-retrieved as relevant to \(q\). The goal is to enable an LLM to generate the corresponding SQL query. In online Text-to-SQL services, user queries are processed in batches to enhance throughput. Importantly, a significant number of queries often access a shared subset of tables—an observation exploited in this work to further optimize the efficiency of Text-to-SQL inference systems.


\section{Method}

\newcommand{\vpara}[1]{\vspace{1.5ex}\noindent\textbf{#1}}

TableCache accelerates inference by maximizing the reuse of precomputed table representations. As shown in Figure~\ref{fig:main}, we precompute KV caches based on the primary-foreign key graph, organizing them into a hierarchical Table Trie offline. During inference, the system employs KV cache management strategies and reranks batched queries to minimize switching overhead. Additionally, a computation pipeline overlaps memory access with processing to further reduce latency.

\subsection{Offline KV Cache Precomputation} 
 

During the offline KV cache precomputation phase, TableCache aims to construct table-level KV cache information for practical inference usage. Specifically, we independently compute and store KV caches for individual tables, which are subsequently concatenated during inference. This process transforms the standard global causal attention mask into a series of isolated lower-triangular blocks, inevitably severing inter-block attention dependencies and leading to performance degradation.
To address this, we introduce a \textbf{Primary Foreign Key Guided Table Representation} that leverages schema links to reconstruct inter-table attention. For efficient inference, TableCache organizes these precomputed caches into a memory-resident \textbf{Table Trie}, enabling rapid retrieval of necessary KV caches through structural matching.

\begin{figure}
  \centering
  \includegraphics[width=1\linewidth]{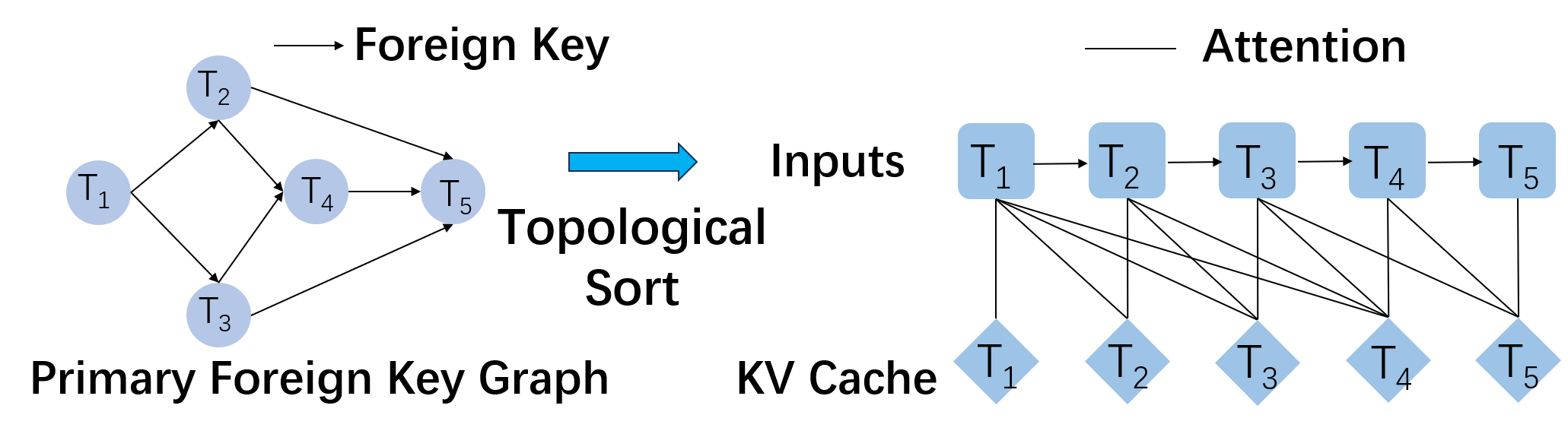}
  \caption{Primary Foreign Key Guided Table Representation.}
  \label{fig:er}
\end{figure}

\subsubsection{Primary Foreign Key Guided Table Representation.} 



As shown in Figure~\ref{fig:er}, we leverage the inherent characteristics of Text-to-SQL tasks to facilitate primary foreign key guided table representation. Specifically, foreign keys often represent relationships between tables in a database. Therefore, if attention between tables linked by foreign keys is lost during attention modeling, it can increase model confusion. This insight inspired us to construct a primary foreign key graph of the original database. Specifically, we first abstract the tables in the database as nodes(treating isolated ones as zero-degree) in the graph, and then for each table, we create edges from the tables referenced by its foreign keys to itself. Leveraging the inherent acyclic nature of standard schemas, we apply topological sorting on this graph, ensuring that nodes later in the topological order necessarily depend on preceding nodes. This dependency allows connected nodes to be jointly encoded offline. Through this joint encoding approach, we restore the most essential attention relationships between tables at minimal cost.


Moreover, it is noteworthy that, consistent with findings from prior studies~\citep{lu2024turborag,ma2024block}, the proper implementation of positional encoding is crucial for maintaining the performance of independently encoded KV cache. Therefore, during the offline precomputation phase, TableCache first stores the KV cache information without positional encoding. Subsequently, during the inference phase, TableCache dynamically re-applies positional encodings to each token based on its global position within the concatenated sequence.

\begin{algorithm}[t]
\caption{Table Trie Matching}
\label{alg:trie}
\LinesNumbered 
\SetNlSty{}{}{:} 
\SetNlSkip{1em} 
\KwIn{$s$: input token sequence}
\KwIn{$trie$: pre-built Table Trie}
\KwIn{$cache$: KV cache manager}
\SetKwFunction{Match}{Match}
\SetKwProg{Fn}{Function}{:}{}
\Fn{\Match{$T$, $trie$, $cache$}}{
    $p \gets 0$\tcp*{current start position}
    $H \gets \emptyset$\;
    \While{$p < l_s$}{
        $(found, next, pos) \gets trie.query(T, p)$\;
        \If{$found$}{
            $kv \gets cache.get(pos)$\;
            $H.add(kv)$\;
            $p \gets next$\tcp*{Next table's start position}
        }
        \Else{
            $p \gets p + 1$\;
        }
    }
    \Return $H$\;
}
\end{algorithm}

\subsubsection{Table Trie Construction}
After constructing the primary foreign key guided table representation, all table information is organized into a trie structure called \textbf{Table Trie}. Specifically, the contents (i.e., table name, column names/descriptions) of each table are inserted into the Table Trie independently. The leaf nodes of the Table Trie store the table ID and storage path to its corresponding KV cache. 

After the Table Trie is built, during the inference stage, the system identifies all tables in a given query by performing continuous matching within the Table Trie. As shown in the Algorithm~\ref{alg:trie}, the system finds the longest prefix match between the query and the Table Trie (Line-5) for each match and retrieves the corresponding KV cache from the matched node (Line-7). The matched prefix is then removed from the query (Line-9), and the process is repeated until no further matches can be found. All relevant tables associated with the query are collected through this iterative approach. Finally, the KV caches of these tables are concatenated and fed into the LLM for subsequent inference. 

\subsection{Table KV Cache Management}


Since TableCache precomputes KV caches offline, the prefilling bottleneck shifts from computation to memory access. To address this, we designed a KV cache management system using GPU eviction strategies (FIFO, LRU, and LFU) to efficiently identify query hotspots. This system triggers updates exclusively when a request targets a non-cached table, necessitating a swap-in operation where the eviction candidate is selected by the cache management module. Specific implementations are detailed in Appendix~\ref{sec:manager}.

However, in practice, the order of user queries significantly impacts cache eviction frequency. As shown in Figure~\ref{fig:main}(a), a cache-unfriendly sequence causes frequent, redundant replacements (e.g., repeatedly evicting and reloading Tables 1–6). In contrast, reordering the sequence loads these tables only once. Since high switching overhead severely degrades performance, we introduce a query reranking strategy to reorganize input batches into memory-friendly sequences. As evidenced by our experimental results (Table~\ref{tab:manager}), it is precisely this reranking mechanism that empowers simple cache eviction policies to effectively capture significant query hotspots.


To maximize resource efficiency, TableCache employs a computation loading pipeline that prefetches KV caches concurrently with GPU computation. This strategy overlaps memory access with processing, effectively hiding latency and reducing overall response time. We introduce the details of the query reranking and computational loading pipeline below.

\subsubsection{Query Reranking} 

The objective of Query Reranking is to minimize the frequency of KV cache switches. To this end, we employ a greedy strategy that iteratively selects from the remaining queries based on maximal table similarity to the immediate predecessor, starting from a random anchor.

To define the similarity between queries, we first match each query against the Table Trie to identify the associated set of table IDs, denoted as $\tau_i$. Next, we define a binary array $I_i$ of length $n$ (where $n$ represents the total number of tables across all given databases) for query $i$, where each

\begin{equation}
\label{eq:bin}
    I_i[k] = 
    \begin{cases} 
        1, & \text{if } k \in \tau_i\\
        0, & \text{if } k \notin \tau_i \\
    \end{cases}.
\end{equation}

Next, we define the distance between two binary strings \(I_i\) and \(I_j\) as $d_{i,j}$.

\begin{equation}
\label{eq:dis}
    d_{i,j} = 
    \sum_{k=1}^n (I_i \oplus I_j)[k].
\end{equation}

\noindent where $\oplus$ denotes the binary XOR operation. This distance metric quantifies the difference in the sets of tables associated with query $i$ and query $j$, thereby measuring their similarity. Based on this, at each step, we select the query with the minimal distance relative to the current reference point. 


\subsubsection{Computation Loading Pipeline} In practice, the inference process of LLMs can primarily be decomposed into two stages: computation and loading. Since TableCache pre-computes the KV cache for each table, the computation and cache loading for the TableCache can proceed in parallel. Subsequently, TableCache divides the $bs$ queries in a batch into smaller micro-batches of size $b_c$ and $b_m$, where $b_c$ represents the compute micro-batch and $b_m$ the memory micro-batch. During inference, TableCache computes the current $b_c$ requests while simultaneously prefetching the table information and KV cache required for the next $b_m$ requests onto the GPU, thus achieving pipeline parallelism.

Compared to other KV cache management approaches, our solution fundamentally decouples computation from memory access during inference. In contrast, conventional methods rely on real-time GPU computation to obtain the KV cache, thereby making prefetching impossible and consequently failing to maximize memory performance. By leveraging this parallelization, we maximize the utilization of both GPU and CPU resources, achieving significant acceleration for the prefill phase.

\vpara{Discussion.} Crucially, integrating the above modules into the whole cache management framework is straightforward and incurs negligible overhead. The reranking step occurs just once per batch prior to inference, while the pipeline simply performs data prefetching synchronously during model execution without complex control logic. This ensures that the system remains practically feasible without compromising efficiency.



\begin{table*}[h]
\centering
\renewcommand{\arraystretch}{0.8}
\setlength{\tabcolsep}{3.0mm}{
\begin{tabular}{l|c|cc|cc}
\toprule
\multirow{2}{*}{Method} & \multirow{2}{*}{Model} & \multicolumn{2}{c|}{Spider\textsubscript{dev}} & \multicolumn{2}{c}{BIRD\textsubscript{dev}} \\
\cmidrule{3-6}
                        &                        & TTFT↓ (s) & Acc.↑ (\%) & TTFT↓ (s) & Acc.↑ (\%) \\
\midrule
Baseline                & \multirow{4}{*}{\makecell[c]{OmniSQL-7B}} & 78.30    & 78.3      & 296.42   & 61.5      \\
vLLM                    &                        & 68.72    & 77.5      & 177.36   & 61.5      \\
SGLang                  &                        & 72.23    & 78.6      & 199.85   & 61.3      \\
TableCache              &                        & \textbf{42.43}        & 77.1        & \textbf{90.39}        & 59.9      \\
\midrule
Baseline                & \multirow{4}{*}{\makecell[c]{Qwen2.5-7B-coder}} & 74.81    & 73.2      & 301.35   & 54.3      \\
vLLM                    &                        & 66.08    & 71.5      & 207.49   & 54.1      \\
SGLang                  &                        & 62.16    & 72.3      & 177.43   & 54.1      \\
TableCache              &                        & \textbf{41.27}        & 72.2        & \textbf{117.22}        & 53.3      \\
\bottomrule
\end{tabular}
}
\caption{The results of the comparison methods presented on two backbone models across two benchmarks. The baseline represents direct inference using the Transformers library, while the results for both vLLM and SGLang are obtained with radixcache enabled. Boldface indicates the best performance.}
\label{tab:main_result}
\end{table*}

\section{Experiments}

\subsection{Experiment Setup}

\vpara{Datasets.} To evaluate the effectiveness of our method, we conducted experiments on two text-to-SQL benchmarks: BIRD~\citep{li2023can}and Spider~\citep{yu2018spider}.

\vpara{BIRD}~\citep{li2023can} is a Text-to-SQL benchmark comprising a total of 12,751 questions, spanning 37 different specialized domains, emphasizing the model’s ability to integrate and handle a wide range of external knowledge.

\vpara{Spider}~\citep{li2023can} is a large-scale, cross-domain Text-to-SQL dataset designed to assess the generalization capabilities across various databases and query structures.


To better simulate a realistic online Text-to-SQL scenario, we randomized the order of data samples in the dataset as well as the table order provided for each sample.

\vpara{Baselines.} We compare our method against a standard Baseline (direct inference without acceleration) and two categories of optimized systems. First, for mainstream inference engines, we evaluate SGLang~\citep{zheng2025sglang} and vLLM~\citep{kwon2023efficient}, both configured with RadixCache. Second, for block-cache methods, we select PromptCache~\citep{gim2024prompt} and TurboRAG~\citep{lu2024turborag}, as they natively support the Qwen model family.

\vpara{Metrics.} Following previous research~\citep{lu2024turborag,chan2024dontragcacheaugmentedgeneration,yao2024cacheblendfastlargelanguage}, we primarily adopted \textbf{t}otal delay \textbf{t}ime to \textbf{f}irst \textbf{t}oken (TTFT/s) across the entire test set and the Execution Accuracy (\textbf{Acc.}) of the generated SQL statements as our metrics.

\begin{table*}[t]
\centering
\setlength{\tabcolsep}{3.3mm}
\begin{tabular}{lcccc}
\toprule
\textbf{Dataset} & \textbf{w/o Cache Mgmt.} & \textbf{w/o Query Rerank} & \textbf{w/o Comp. Load.} & \textbf{TableCache} \\
\midrule
Spider\textsubscript{dev} & 99.346 & 58.333 & 38.549 & \textbf{36.229} \\
Bird\textsubscript{dev} & 212.626 & 114.529 & 86.842 & \textbf{80.526} \\
\bottomrule
\end{tabular}
\caption{Ablation studies on cache management, query ranking, and compute loading pipelines on Spider\textsubscript{dev} and BIRD\textsubscript{dev}, using OmniSQL-7B, with TTFT measured in seconds.}
\label{tab:ablation}
\end{table*}

\subsection{Implementation Details}


For each benchmark, we extracted referenced tables to pre-compute KV caches and organized them into the offline Table Trie. Using OmniSQL-7B~\citep{li2025omnisql} and Qwen2.5-7B-Coder~\citep{qwen2.5} as representative backbones, we implemented an adaptive training strategy to align them with the Primary Foreign Key Guided Table Representation.

Offline KV caches introduce block-wise attention by preserving intra-table and inter-table token attention for tables connected via primary foreign key relationships. This contrasts with the original full attention KV caches, which span all input table tokens. To adapt backbone models previously trained on full attention to TableCache’s block-wise KV caches, we introduced an attention mask in the training process to restrict attention to tokens within individual tables\footnote{As KV caching is unavailable during training, we omitted PFK-based inter-table attention to streamline implementation. Both independent table chunking and PFK-guided chunking manifest as series of lower-triangular attention blocks—differing only in block size. Due to this formal identity, the simplified training mask is sufficient to adapt the model to TableCache’s inference-time attention patterns.}. For fair comparison, the backbones used with the other baselines were also trained, but without the specialized attention mask.

To adapt the backbones, we fine-tuned them on complex tables from $\text{BIRD}_{\text{train}}$ for three epochs (see Appendix~\ref{sec:tr}). Experiments were conducted on a single NVIDIA A800 GPU with pipeline parameters $b_c = 100$ and $b_m = 10$. The inference prompt is provided in Appendix~\ref{sec:prompt}.

\subsection{Main Results}
Table~\ref{tab:main_result} presents the following findings:

\vpara{TableCache Achieves Outstanding Latency Optimization.} In the experiments conducted on OmniSQL-7B and Qwen2.5-7B-coder, our proposed TableCache method significantly outperformed the existing baseline. The experimental results demonstrate that, across two distinct benchmarks, our method achieves up to a \(2.69\times\) and \(3.62\times\) reduction in TTFT compared to directly using the transformers library for inferecne (i.e, Baseline in Table~\ref{tab:main_result}). Moreover, our method demonstrates an even higher acceleration ratio on BIRD, a longer and more challenging benchmark compared to Spider, surpassing inference engines like vLLM and SGLang that employ CUDA kernel level optimizations. This indicates that in Text-to-SQL scenarios involving multiple tables that may be shared across users, with potentially random table orders, the proposed TableCache is more adaptable to these scenarios, effectively reducing inference latency during the prefill phase.

\vpara{TableCache Achieves Near-Lossless Performance in Text-to-SQL Models.} We find that the integration of TableCache into existing models results in no significant performance loss (accuracy gaps within 1\%). This success stems from our Primary Foreign Key Guided Table Representation, which effectively recovers and stores inter-block dependencies that are otherwise compromised by block-wise encoding. Consequently, TableCache guarantees high acceleration through block-wise encoding while maintaining accuracy.




\begin{table}[t]
\centering
\begin{tabular}{lcc}
\toprule
\textbf{Dataset} & \textbf{w/o PFTR} & \textbf{TableCache} \\
\midrule
Spider & 72.0 & \textbf{76.9} \\
Bird & 50.9 & \textbf{58.1} \\
\bottomrule
\end{tabular}
\caption{Ablation study on the primary foreign key guided table representation, reported in terms of Accuracy. The backbone model is OmniSQL-7B(zero-shot). PFTR refers to \underline{P}rimary \underline{F}oreign key guided \underline{T}able \underline{R}epresentation.}
\label{tab:caching_results}
\end{table}

\subsection{Ablation Study}




To further analyze the effectiveness of the proposed modules, we conducted an ablation study on the individual components that constitute TableCache, and the results are presented in Table~\ref{tab:ablation} and Table~\ref{tab:caching_results}. 

\vpara{Ablation of Primary Foreign Key Guided Table Representation.} Table~\ref{tab:caching_results} shows that failing to reconstruct the relationships between tables with primary foreign key connections leads to significant performance degradation on both datasets, with an even more pronounced drop observed on the more challenging BIRD dataset. This result further supports our assertion that, in real-world database scenarios, inter-table relationships are primarily governed by foreign key joins—relationships that are indispensable for solving complex Text-to-SQL tasks. A typical example illustrating the importance of primary foreign key guided table representation is provided in Appendix~\ref{sec:case}.

\begin{table}[t]
    \centering
    \setlength{\tabcolsep}{6mm}{
    \begin{tabular}{ll}
        \toprule
        \textbf{Cache Eviction Strategy} & \textbf{TTFT (s)} \\
        \midrule
        LFU  & 36.85 \\
        FIFO & 36.41 \\
        LRU  & \textbf{36.22} \\
        \bottomrule
    \end{tabular}
    }
    \caption{Ablation studies on cache eviction strategies on \(\text{Spider}_{\text{dev}}\) using OmniSQL-7B.}
    \label{tab:manager}
\end{table}

\vpara{Ablation of Table KV Cache Management.} As shown in Table~\ref{tab:ablation}, the absence of this component leads to a drastic increase in TTFT. This confirms that without the cache management system, the naive approach of repeatedly loading caches from CPU to GPU and immediately offloading them creates a severe I/O bottleneck, ultimately crippling system performance.

Ablation results on $\text{Spider}_{\text{dev}}$ (Table~\ref{tab:manager}) show that caching strategies (LRU, LFU, FIFO) have minimal impact on prefill efficiency. This is because hotspot patterns are easily captured, and query reranking concentrates hot data in short time windows, minimizing eviction.

\begin{table*}[h]
\centering
\renewcommand{\arraystretch}{0.8}
\setlength{\tabcolsep}{3.0mm}{
\begin{tabular}{l|c|cc|cc}
\toprule
\multirow{2}{*}{Method} & \multirow{2}{*}{Model} & \multicolumn{2}{c|}{Spider\textsubscript{dev}} & \multicolumn{2}{c}{BIRD\textsubscript{dev}} \\
\cmidrule{3-6}
                        &                        & TTFT↓ (s) & Acc.↑ (\%) & TTFT↓ (s) & Acc.↑ (\%) \\
\midrule
Baseline                & \multirow{6}{*}{OmniSQL-7B} & 97.51   & 78.3      & 291.26   & 61.2      \\
vLLM                    &                        & 61.17   & 77.4      & 159.88   & 60.4      \\
SGLang                  &                        & 53.16   & 77.9      & 128.49   & 61.0      \\
PromptCache             &                        & 55.47   & 66.2      & 85.01    & 49.7      \\
TurboRAG                &                        & 81.47        & 72.1        & 205.43        & 51.0         \\
TableCache              &                        & \textbf{36.22} & 76.9 & \textbf{80.53} & 58.1      \\
\midrule
Baseline                & \multirow{6}{*}{Qwen2.5-7B-coder} & 98.13   & 73.5      & 294.83   & 51.0      \\
vLLM                    &                        & 75.38   & 73.0      & 233.74   & 48.2      \\
SGLang                  &                        & 73.21   & 72.4      & 223.58   & 50.5      \\
PromptCache             &                        & 52.36        & 51.4        & 133.47        & 25.4      \\
TurboRAG                &                        & 77.40        & 56.0       & 203.11        & 29.2         \\
TableCache              &                        & \textbf{42.01} & 72.0 & \textbf{115.36} & 42.9      \\
\bottomrule
\end{tabular}
}
\caption{Ablation study on the training process and comparison with additional training-free and block-wise methods. Boldface indicates the best performance.}
\label{tab:wotrain}
\end{table*}

\vpara{Ablation of Query Reranking.} The objective of Query Reranking is to minimize memory exchanges between the CPU and GPU. As shown in Table~\ref{tab:ablation}, disabling reranking significantly increases the system's TTFT. This occurs because disordered user inputs disrupt table access patterns, making it challenging for cache eviction policies to identify hot data. As a result, excessive swapping operations are required, leading to a notable decline in overall system efficiency.


\vpara{Ablation of Computation Loading Pipeline.} Unlike modules focused on reducing swap frequency, this module leverages TableCache's KV precomputation to decouple computation from memory access. As shown in Table~\ref{tab:ablation}, disabling this pipeline increases TTFT, demonstrating that the module effectively enhances performance by masking memory latency within computation.

\vpara{Ablation of Model Training} Table~\ref{tab:wotrain} shows TableCache's performance in a training-free setting. While TableCache exhibits robust results on Spider, it achieves lossless performance on pre-trained Text-to-SQL models like OmniSQL even without further tuning. For non-specialized models like Qwen2.5-7B-coder, TableCache incurs minor degradation but still significantly surpasses existing training-free, block-wise encoding methods. Notably, TableCache reaches SOTA efficiency in TTFT reduction.



\section{Discussion}
\subsection{Why TableCache Works?}
TableCache achieves superior performance by leveraging two domain characteristics. First, we exploit the static, relation-rich nature of schemas to enable offline pre-computation and PFK representations. Second, we utilize the temporal locality of user queries to drive our reranking and pipelining strategies, effectively addressing Text-to-SQL bottlenecks that generic caching cannot.

\subsection{Practical Feasibility}
First, in terms of applicability, the inherent dependency of Text-to-SQL on schema information aligns our method with enterprise scenarios where data permissions are established. Second, concerning overhead, the static nature of schemas in these settings implies that precomputation is a one-time cost. Because this cost is amortized over long-term usage, it remains highly efficient compared to the recurring latency of online processing. Furthermore, as such scenarios often require model fine-tuning for domain adaptation, TableCache’s adaptive training can be seamlessly merged into the existing optimization workflow.

\subsection{System Time Complexity}
TableCache consists of the Primary Foreign Key Guided Table Representation, Table Trie, Query Reranking, and Computation Loading Pipeline. Defining $m$ as the table count, $n$ as the user input length, $q$ as the query length, and $N$ as the number of queries, the complexities are as follows: the representation construction is $O(m)$, Table Trie matching is $O(n)$, and query reranking is $O(\frac{N^2m}{64})$. The overall system inference complexity is $O(nq)$. A detailed complexity analysis is provided in Appendix~\ref{sec:comp}.
\section{Conclusion}
This paper highlights that in text-to-SQL scenarios, current inference engines like SGLang and vLLM fail to efficiently utilize KV caches from frequently queried tables, resulting in substantial prefill overhead and increased request latency. To address this issue, we propose TableCache, which optimizes performance by reusing hot-table KV caches—precomputed table representations guided by primary foreign key graphs—and integrating a custom cache management system with a query ranking mechanism and computation loading pipeline. TableCache significantly reduces Time to First Token (TTFT) while preserving model accuracy. Experimental results demonstrate that TableCache achieves the lowest TTFT across multiple text-to-SQL benchmarks.
\section{Limitations}

While TableCache is theoretically extensible to broader tasks like textual QA and KBQA, we confined this work to the Text-to-SQL scenario. This choice is motivated by the domain's unique advantages: explicit foreign key constraints provide stronger structural signals than the implicit relationships found in general text, and table granularity offers a well-defined retrieval boundary. Extending to unstructured domains requires addressing these complex, implicit dependencies, which we leave for future research—potentially by integrating with graph retrieval methods. Additionally, the modules developed herein are designed for high interoperability and can be seamlessly incorporated into existing KV-cache management systems, such as vLLM and SGLang, to further enhance performance.
\bibliography{custom}

\begin{thebibliography}{29}
\providecommand{\natexlab}[1]{#1}

\bibitem[{Cai et~al.(2024)Cai, Zhang, Gao, Liu, Liu, Lu, Xiong, Dong, Chang, Hu, and Xiao}]{Cai2024PyramidKVDK}
Zefan Cai, Yichi Zhang, Bofei Gao, Yuliang Liu, Tianyu Liu, Keming Lu, Wayne Xiong, Yue Dong, Baobao Chang, Junjie Hu, and Wen Xiao. 2024.
\newblock \href {https://api.semanticscholar.org/CorpusID:270226243} {Pyramidkv: Dynamic kv cache compression based on pyramidal information funneling}.
\newblock \emph{ArXiv}, abs/2406.02069.

\bibitem[{Chan et~al.(2024)Chan, Chen, Cheng, and Huang}]{chan2024dontragcacheaugmentedgeneration}
Brian~J Chan, Chao-Ting Chen, Jui-Hung Cheng, and Hen-Hsen Huang. 2024.
\newblock \href {https://arxiv.org/abs/2412.15605} {Don't do rag: When cache-augmented generation is all you need for knowledge tasks}.
\newblock \emph{Preprint}, arXiv:2412.15605.

\bibitem[{Cheng et~al.(2023)Cheng, Xie, Shi, Li, Nadkarni, Hu, Xiong, Radev, Ostendorf, Zettlemoyer et~al.}]{cheng2023binding}
Zhoujun Cheng, Tianbao Xie, Peng Shi, Chengzu Li, Rahul Nadkarni, Yushi Hu, Caiming Xiong, Dragomir Radev, Mari Ostendorf, Luke Zettlemoyer, and 1 others. 2023.
\newblock Binding language models in symbolic languages.
\newblock In \emph{International Conference on Learning Representations (ICLR 2023)(01/05/2023-05/05/2023, Kigali, Rwanda)}.

\bibitem[{Feng et~al.(2024)Feng, Lv, Cao, Xie, and Zhou}]{feng2024ada}
Yuan Feng, Junlin Lv, Yukun Cao, Xike Xie, and S~Kevin Zhou. 2024.
\newblock Ada-kv: Optimizing kv cache eviction by adaptive budget allocation for efficient llm inference.
\newblock \emph{arXiv preprint arXiv:2407.11550}.

\bibitem[{Gao et~al.(2024{\natexlab{a}})Gao, Wang, Li, Sun, Qian, Ding, and Zhou}]{gao2024text}
Dawei Gao, Haibin Wang, Yaliang Li, Xiuyu Sun, Yichen Qian, Bolin Ding, and Jingren Zhou. 2024{\natexlab{a}}.
\newblock Text-to-sql empowered by large language models: A benchmark evaluation.
\newblock \emph{Proceedings of the VLDB Endowment}, 17(5):1132--1145.

\bibitem[{Gao et~al.(2024{\natexlab{b}})Gao, Liu, Li, Shi, Zhu, Wang, Li, Li, Hong, Luo et~al.}]{gao2024xiyan}
Yingqi Gao, Yifu Liu, Xiaoxia Li, Xiaorong Shi, Yin Zhu, Yiming Wang, Shiqi Li, Wei Li, Yuntao Hong, Zhiling Luo, and 1 others. 2024{\natexlab{b}}.
\newblock Xiyan-sql: A multi-generator ensemble framework for text-to-sql.
\newblock \emph{arXiv preprint arXiv:2411.08599}.

\bibitem[{Gim et~al.(2024)Gim, Chen, Lee, Sarda, Khandelwal, and Zhong}]{gim2024prompt}
In~Gim, Guojun Chen, Seung-seob Lee, Nikhil Sarda, Anurag Khandelwal, and Lin Zhong. 2024.
\newblock Prompt cache: Modular attention reuse for low-latency inference.
\newblock \emph{Proceedings of Machine Learning and Systems}, 6:325--338.

\bibitem[{Gim et~al.(2023)Gim, Chen, seob Lee, Sarda, Khandelwal, and Zhong}]{Gim2023PromptCM}
In~Gim, Guojun Chen, Seung seob Lee, Nikhil Sarda, Anurag Khandelwal, and Lin Zhong. 2023.
\newblock \href {https://api.semanticscholar.org/CorpusID:265067391} {Prompt cache: Modular attention reuse for low-latency inference}.
\newblock \emph{ArXiv}, abs/2311.04934.

\bibitem[{Hu et~al.(2024)Hu, Huang, Wang, Wang, Hu, Zhang, Feng, Chen, Shan, and Xie}]{hu2024epic}
Junhao Hu, Wenrui Huang, Weidong Wang, Haoyi Wang, Tiancheng Hu, Qin Zhang, Hao Feng, Xusheng Chen, Yizhou Shan, and Tao Xie. 2024.
\newblock Epic: Efficient position-independent caching for serving large language models.
\newblock \emph{arXiv preprint arXiv:2410.15332}.

\bibitem[{Kim et~al.(2025)Kim, Kim, Kwon, Lee, Yun, and Song}]{kim2025kvzip}
Jang-Hyun Kim, Jinuk Kim, Sangwoo Kwon, Jae~W Lee, Sangdoo Yun, and Hyun~Oh Song. 2025.
\newblock Kvzip: Query-agnostic kv cache compression with context reconstruction.
\newblock \emph{arXiv preprint arXiv:2505.23416}.

\bibitem[{Kwon et~al.(2023)Kwon, Li, Zhuang, Sheng, Zheng, Yu, Gonzalez, Zhang, and Stoica}]{kwon2023efficient}
Woosuk Kwon, Zhuohan Li, Siyuan Zhuang, Ying Sheng, Lianmin Zheng, Cody~Hao Yu, Joseph Gonzalez, Hao Zhang, and Ion Stoica. 2023.
\newblock Efficient memory management for large language model serving with pagedattention.
\newblock In \emph{Proceedings of the 29th Symposium on Operating Systems Principles}, pages 611--626.

\bibitem[{Li et~al.(2025)Li, Wu, Zhang, Huang, Zhang, Jiang, Wang, Zhang, Chen, Shi et~al.}]{li2025omnisql}
Haoyang Li, Shang Wu, Xiaokang Zhang, Xinmei Huang, Jing Zhang, Fuxin Jiang, Shuai Wang, Tieying Zhang, Jianjun Chen, Rui Shi, and 1 others. 2025.
\newblock Omnisql: Synthesizing high-quality text-to-sql data at scale.
\newblock \emph{arXiv preprint arXiv:2503.02240}.

\bibitem[{Li et~al.(2024{\natexlab{a}})Li, Zhang, Liu, Fan, Zhang, Zhu, Wei, Pan, Li, and Chen}]{li2024codes}
Haoyang Li, Jing Zhang, Hanbing Liu, Ju~Fan, Xiaokang Zhang, Jun Zhu, Renjie Wei, Hongyan Pan, Cuiping Li, and Hong Chen. 2024{\natexlab{a}}.
\newblock Codes: Towards building open-source language models for text-to-sql.
\newblock \emph{Proceedings of the ACM on Management of Data}, 2(3):1--28.

\bibitem[{Li et~al.(2023)Li, Hui, Qu, Yang, Li, Li, Wang, Qin, Geng, Huo et~al.}]{li2023can}
Jinyang Li, Binyuan Hui, Ge~Qu, Jiaxi Yang, Binhua Li, Bowen Li, Bailin Wang, Bowen Qin, Ruiying Geng, Nan Huo, and 1 others. 2023.
\newblock Can llm already serve as a database interface? a big bench for large-scale database grounded text-to-sqls.
\newblock \emph{Advances in Neural Information Processing Systems}, 36:42330--42357.

\bibitem[{Li et~al.(2024{\natexlab{b}})Li, Huang, Yang, Venkitesh, Locatelli, Ye, Cai, Lewis, and Chen}]{li2024snapkv}
Yuhong Li, Yingbing Huang, Bowen Yang, Bharat Venkitesh, Acyr Locatelli, Hanchen Ye, Tianle Cai, Patrick Lewis, and Deming Chen. 2024{\natexlab{b}}.
\newblock Snapkv: Llm knows what you are looking for before generation.
\newblock \emph{Advances in Neural Information Processing Systems}, 37:22947--22970.

\bibitem[{Lu et~al.(2024)Lu, Wang, Rong, Chen, and Tang}]{lu2024turborag}
Songshuo Lu, Hua Wang, Yutian Rong, Zhi Chen, and Yaohua Tang. 2024.
\newblock Turborag: Accelerating retrieval-augmented generation with precomputed kv caches for chunked text.
\newblock \emph{arXiv preprint arXiv:2410.07590}.

\bibitem[{Ma et~al.(2024)Ma, Wang, and Tian}]{ma2024block}
Dongyang Ma, Yan Wang, and Lan Tian. 2024.
\newblock Block-attention for efficient prefilling.
\newblock \emph{arXiv preprint arXiv:2409.15355}.

\bibitem[{Nan et~al.(2023)Nan, Zhao, Zou, Ri, Tae, Zhang, Cohan, and Radev}]{nan2023enhancing}
Linyong Nan, Yilun Zhao, Weijin Zou, Narutatsu Ri, Jaesung Tae, Ellen Zhang, Arman Cohan, and Dragomir Radev. 2023.
\newblock Enhancing few-shot text-to-sql capabilities of large language models: A study on prompt design strategies.
\newblock \emph{arXiv preprint arXiv:2305.12586}.

\bibitem[{Pourreza and Rafiei(2023)}]{pourreza2023din}
Mohammadreza Pourreza and Davood Rafiei. 2023.
\newblock Din-sql: decomposed in-context learning of text-to-sql with self-correction.
\newblock In \emph{Proceedings of the 37th International Conference on Neural Information Processing Systems}, pages 36339--36348.

\bibitem[{Pourreza and Rafiei(2024)}]{pourreza2024dts}
Mohammadreza Pourreza and Davood Rafiei. 2024.
\newblock Dts-sql: Decomposed text-to-sql with small large language models.
\newblock In \emph{EMNLP (Findings)}.

\bibitem[{Talaei et~al.(2024)Talaei, Pourreza, Chang, Mirhoseini, and Saberi}]{talaei2024chess}
Shayan Talaei, Mohammadreza Pourreza, Yu-Chen Chang, Azalia Mirhoseini, and Amin Saberi. 2024.
\newblock Chess: Contextual harnessing for efficient sql synthesis.
\newblock \emph{arXiv preprint arXiv:2405.16755}.

\bibitem[{Tang et~al.(2024)Tang, Zhao, Zhu, Xiao, Kasikci, and Han}]{tang2024quest}
Jiaming Tang, Yilong Zhao, Kan Zhu, Guangxuan Xiao, Baris Kasikci, and Song Han. 2024.
\newblock Quest: Query-aware sparsity for efficient long-context llm inference.
\newblock \emph{arXiv preprint arXiv:2406.10774}.

\bibitem[{Yang et~al.(2024)Yang, Yang, Zhang, Hui, Zheng, Yu, Li, Liu, Huang, Wei, Lin, Yang, Tu, Zhang, Yang, Yang, Zhou, Lin, Dang, Lu, Bao, Yang, Yu, Li, Xue, Zhang, Zhu, Men, Lin, Li, Xia, Ren, Ren, Fan, Su, Zhang, Wan, Liu, Cui, Zhang, and Qiu}]{qwen2.5}
An~Yang, Baosong Yang, Beichen Zhang, Binyuan Hui, Bo~Zheng, Bowen Yu, Chengyuan Li, Dayiheng Liu, Fei Huang, Haoran Wei, Huan Lin, Jian Yang, Jianhong Tu, Jianwei Zhang, Jianxin Yang, Jiaxi Yang, Jingren Zhou, Junyang Lin, Kai Dang, and 22 others. 2024.
\newblock Qwen2.5 technical report.
\newblock \emph{arXiv preprint arXiv:2412.15115}.

\bibitem[{Yang et~al.(2025)Yang, Hou, Wei, Bao, and Chang}]{yang2025kvlink}
Jingbo Yang, Bairu Hou, Wei Wei, Yujia Bao, and Shiyu Chang. 2025.
\newblock Kvlink: Accelerating large language models via efficient kv cache reuse.
\newblock \emph{arXiv preprint arXiv:2502.16002}.

\bibitem[{Yao et~al.(2024)Yao, Li, Liu, Ray, Cheng, Zhang, Du, Lu, and Jiang}]{yao2024cacheblendfastlargelanguage}
Jiayi Yao, Hanchen Li, Yuhan Liu, Siddhant Ray, Yihua Cheng, Qizheng Zhang, Kuntai Du, Shan Lu, and Junchen Jiang. 2024.
\newblock \href {https://arxiv.org/abs/2405.16444} {Cacheblend: Fast large language model serving for rag with cached knowledge fusion}.
\newblock \emph{Preprint}, arXiv:2405.16444.

\bibitem[{Yu et~al.(2018)Yu, Zhang, Yang, Yasunaga, Wang, Li, Ma, Li, Yao, Roman et~al.}]{yu2018spider}
Tao Yu, Rui Zhang, Kai Yang, Michihiro Yasunaga, Dongxu Wang, Zifan Li, James Ma, Irene Li, Qingning Yao, Shanelle Roman, and 1 others. 2018.
\newblock Spider: A large-scale human-labeled dataset for complex and cross-domain semantic parsing and text-to-sql task.
\newblock \emph{arXiv preprint arXiv:1809.08887}.

\bibitem[{Zhang et~al.(2023)Zhang, Sheng, Zhou, Chen, Zheng, Cai, Song, Tian, R{\'e}, Barrett, Wang, and Chen}]{Zhang2023H2OHO}
Zhenyu~(Allen) Zhang, Ying Sheng, Tianyi Zhou, Tianlong Chen, Lianmin Zheng, Ruisi Cai, Zhao Song, Yuandong Tian, Christopher R{\'e}, Clark~W. Barrett, Zhangyang Wang, and Beidi Chen. 2023.
\newblock \href {https://api.semanticscholar.org/CorpusID:259263947} {H2o: Heavy-hitter oracle for efficient generative inference of large language models}.
\newblock \emph{ArXiv}, abs/2306.14048.

\bibitem[{Zheng et~al.(2023)Zheng, Yin, Xie, Huang, Sun, Yu, Cao, Kozyrakis, Stoica, Gonzalez et~al.}]{zheng2023efficiently}
Lianmin Zheng, Liangsheng Yin, Zhiqiang Xie, Jeff Huang, Chuyue Sun, Cody\_Hao Yu, Shiyi Cao, Christos Kozyrakis, Ion Stoica, Joseph~E Gonzalez, and 1 others. 2023.
\newblock Efficiently programming large language models using sglang.

\bibitem[{Zheng et~al.(2025)Zheng, Yin, Xie, Sun, Huang, Yu, Cao, Kozyrakis, Stoica, Gonzalez et~al.}]{zheng2025sglang}
Lianmin Zheng, Liangsheng Yin, Zhiqiang Xie, Chuyue~Livia Sun, Jeff Huang, Cody~Hao Yu, Shiyi Cao, Christos Kozyrakis, Ion Stoica, Joseph~E Gonzalez, and 1 others. 2025.
\newblock Sglang: Efficient execution of structured language model programs.
\newblock \emph{Advances in Neural Information Processing Systems}, 37:62557--62583.

\end{thebibliography}

\appendix

\tcbset{
    colframe=black,
    colback=white,
    coltitle=white,
    colbacktitle=black,
    fonttitle=\bfseries,
    center title,
    boxrule=0.5mm,
    toptitle=1mm,
    before=\par\bigskip\noindent,
    after=\par\bigskip,
    arc=1mm,
}

\clearpage

\section{Cache Manager Details}
\label{sec:manager}
\begin{algorithm}[h]
\caption{LRU Cache Replacement}
\KwIn{$cache$: the cache with a fixed capacity $C$}
\KwIn{$request$: the requested table}
\SetKwFunction{Update}{Update}
\SetKwProg{Fn}{Function}{:}{}
\Fn{\Update{$request$}}{
    \If{$request \in cache$}{
        Move $request$ to the \textbf{front}\;
    }
    \Else{
        \If{$|cache| \geq C$}{
            Evict the item at the \textbf{end}\;
        }
        Insert $request$ at the \textbf{front}\;
    }
}
\end{algorithm}

\begin{algorithm}[h]
\caption{FIFO Cache Replacement}
\KwIn{$cache$: a queue with a fixed capacity $C$}
\KwIn{$request$: the requested item}
\SetKwFunction{Update}{Update}
\SetKwProg{Fn}{Function}{:}{}
\Fn{\Update{$request$}}{
    \If{$request \notin cache$}{
        \If{$|cache| \geq C$}{
            Evict the item at the \textbf{head} of the queue\;
        }
        Enqueue $request$ to the \textbf{tail} of the queue\;
    }
}
\end{algorithm}

\begin{algorithm}[h]
\caption{LFU Cache Replacement}
\KwIn{$cache$: a dictionary mapping item to (frequency, timestamp)}
\KwIn{$request$: the requested item}
\KwIn{Current $timestamp$}
\SetKwFunction{Update}{Update}
\SetKwProg{Fn}{Function}{:}{}
\Fn{\Update{$request$}}{
    \If{$request \in cache$}{
        $cache[request].\text{frequency} \leftarrow cache[request].\text{frequency} + 1$\;
        $cache[request].\text{timestamp} \leftarrow timestamp$\;
    }
    \Else{
        \If{$|cache| \geq C$}{
            Find the item $x$ with the \textbf{smallest} $frequency$\;
            \textbf{(Break ties by selecting the smallest $timestamp$)}\;
            Evict $x$ from the $cache$\;
        }
        Insert $cache[request] \leftarrow (1, timestamp)$\;
    }
}
\end{algorithm}






\section{Training Details}
\label{sec:tr}
During training, the TableCache mechanism utilized the entire $\text{BIRD}_{\text{train}}$ dataset. The model was trained for 3 epochs with a learning rate of 1e-6, a warmup ratio of 0.03, and the Adam optimizer. For the training of TableCache, we aimed to enhance the model's adaptation to the Primary Foreign Key Guided Table Representation by familiarizing it with table-level KV cache behavior. To achieve this, for each data sample in the training set, we first extracted all involved tables and segmented the original input based on table positions. For each individual table, we designed a dedicated attention mask to ensure that during computation, it could only attend to its own content and not access information from other tables. Through this training approach, we further enabled the model to adapt to table-level KV cache, thereby improving model performance.

\section{Prompt Template}
\label{sec:prompt}

\begin{tcolorbox}[title=Prompt Template for model inference]
<tables>\\
Task Overview:\\
You are a data science expert. Below, you are provided with a database schema and a natural language question. Your task is to understand the schema and generate a valid SQL query to answer the question.\\
Database Engine:\\
SQLite
\\This schema describes the database's structure, including tables, columns, primary keys, foreign keys, and any relevant relationships or constraints.
\\Question:\\
<Question>\\
Instructions:\\- Make sure you only output the information that is asked in the question. If the question asks for a specific column, make sure to only include that column in the SELECT clause, nothing more.\\- The generated query should return all of the information asked in the question without any missing or extra information.\\
\end{tcolorbox}
\begin{tcolorbox}[]
Before generating the final SQL query, please think through the steps of how to write the query.\\Output Format:
\\In your answer, please enclose the generated SQL query in a code block:\\```sql\\-- Your SQL query\\```\\Take a deep breath and think step by step to find the correct SQL query.
\end{tcolorbox}

\section{Case Study}
\label{sec:case}
In this section, we analyze the limitations of pure block-wise attention—which processes tables independently without primary foreign key guided table representation—through a representative case study, and further demonstrate the effectiveness of our proposed primary-foreign-key-guided table representation method.
\lstset{
    basicstyle=\ttfamily\small,
    breaklines=true,
    frame=single,
    backgroundcolor=\color{gray!10},
    numbers=left,
    numberstyle=\tiny\color{gray}
}

\begin{mdframed}[backgroundcolor=blue!5,linewidth=0]
\noindent\textbf{Question:} Exclusively virtual refers to Virtual = 'F'. How many schools with an average score in Math greater than 400 in the SAT test are exclusively virtual?

\vspace{0.3cm}
\noindent\textbf{TableCache:}
\begin{lstlisting}
SELECT COUNT(s.CDSCode)
FROM schools AS s
INNER JOIN satscores AS ss ON s.CDSCode = ss.cds
WHERE s.Virtual = 'F' AND ss.AvgScrMath > 400;
\end{lstlisting}

\noindent\textbf{Without primary foreign key graph:}
\begin{lstlisting}
SELECT COUNT(*) 
FROM frpm 
WHERE Virtual = 'F' AND AvgScrMath > 400;
\end{lstlisting}

\noindent\textbf{Ground truth:}
\begin{lstlisting}
SELECT COUNT(DISTINCT T2.School) 
FROM satscores AS T1 
INNER JOIN schools AS T2 ON T1.cds = T2.CDSCode 
WHERE T2.Virtual = 'F' AND T1.AvgScrMath > 400;
\end{lstlisting}
\end{mdframed}

From the above example, we can observe that without the primary foreign key guided table representation, although the model correctly comprehends the query content, it fails to capture the relationship between the satscores and schools tables. These two tables are in fact connected by the CDSCode foreign key, but this critical information is entirely lost when using independent block-wise encoding. Consequently, the model erroneously attempts to retrieve required information from the frpm table.

Our dataset analysis reveals that in the absence of primary foreign key guided table representation, a substantial proportion of erroneous model inferences exhibit precisely this type of failure, constituting the predominant category of discrepancy cases. This finding further validates both the motivation behind and effectiveness of our proposed approach.

\section{Time Complexity Analysis}
\label{sec:comp}
\subsection{Primary Foreign Key Guided Table Representation.}

The time complexity of the topological sorting process comprises two components: graph construction and the sorting algorithm itself.

\textbf{Graph Construction:} This phase has a time complexity of $O(m)$, where $m$ denotes the number of tables. Each table is traversed exactly once to extract foreign key dependencies and build the adjacency list.

\textbf{Topological Sort:} The sorting phase also operates in $O(m)$ time, as each node (table) is visited at most once.
\subsection{Table Trie Construction}
Assuming the sequence length is $n$ and the length matched by the Trie each time is $l_m$, each matching operation requires $O(l_m)$ times. Assuming the starting position of the match is $p$, after each match, the pointer marking the end position jumps to $p + l_m$, meaning at most $O(\frac{n}{l_m})$ matches can occur. Therefore, the final amortized time complexity is $O(\frac{n}{l_m}) * O(l_m) = O(n)$. Therefore, by utilizing Table Trie, we achieve efficient table information matching, laying a solid foundation for subsequent inference.
\subsection{Query Reranking}
To ensure computational efficiency, we implement this module using a bitset in C++. The final time complexity of this procedure is $O ( \frac{N^2m}{64})$, where N is the number of queries and M is the number of tables involved. Since for each pair $(i, j)$, the time complexity of computing the $\oplus$ operation is $O(m)$, this can be optimized to $O(\frac{m}{64})$ on a 64-bit machine by utilizing a bitset. Given $N^2$ possible (i,j) pairs, the total time complexity is $O(\frac{N^2m}{64})$, with the division by 64 reflecting the efficiency gained from 64-bit bitset operations.
\subsection{Overall Inference Time}

To analyze the overall inference time complexity, we first examine the complexity of a single request and then that of batched inference. We represent real-world user input as a combination of \textbf{instruction data}, \textbf{schema information}, and a \textbf{user query}. Typically, the query length $q$ is negligible compared to the total input length $n$ (i.e., $q \ll n$).

During actual online inference, the time overhead of TableCache can be decomposed into the following parts:

\textbf{Table Trie matching overhead:} This part incurs an overhead of $O(n)$.

\textbf{Table loading overhead:} Due to our proposed computation loading pipeline, we prefetch the KV cache of tables required in the future during computation. Therefore, this overhead is hidden within the computation cost and can thus be ignored here.

\textbf{Computation overhead:} The overhead is $O(qn)$. This is because only the KV values of the user’s query need to be computed, and it can attend to all preceding table content. 

Therefore, the total time overhead of a single request is $O(n + qn) = O(qn)$. In contrast, the original time overhead is $O(n^2)$. Since $q << n$, our TableCache also achieves an improvement in theoretical time complexity.

Regarding the overall system, the complexity for $N$ requests is $O(Nnq)$; inclusive of the query reranking module's overhead, the total complexity is derived as $O(Nnq + \frac{N^2m}{64})$.
\end{document}